\documentclass[lettersize,journal]{IEEEtran}
\usepackage{amsmath,amsfonts}
\usepackage[ruled]{algorithm2e} 

\usepackage{array}
\usepackage[caption=false,font=normalsize,labelfont=sf,textfont=sf]{subfig}
\usepackage[dvipsnames,svgnames]{xcolor}
\usepackage[pagebackref=true,breaklinks=true,letterpaper=true,colorlinks,bookmarks=false,linkcolor=red,anchorcolor=blue,citecolor=blue]{hyperref}
\usepackage{textcomp}
\usepackage{stfloats}
\usepackage{url}
\usepackage{verbatim}
\usepackage{graphicx}
\usepackage{cite}
\usepackage{orcidlink}
\usepackage{booktabs} 
\usepackage{enumitem}
\usepackage{colortbl}
\usepackage[accsupp]{axessibility}

\usepackage[capitalize]{cleveref}
\crefname{section}{Sec.}{Secs.}
\Crefname{section}{Section}{Sections}
\Crefname{table}{Table}{Tables}
\crefname{table}{Tab.}{Tabs.}

\newcommand{\etal}{\textit{et al.}\ }
\definecolor{MyGray}{rgb}{0.9, 0.9, 0.9}

\begin{document}

\newcommand{\makeorcid}[1]{\hspace{-1mm}{~\orcidlink{#1}}}

\title{VividDreamer: Towards High-Fidelity and Efficient Text-to-3D Generation}

\author{Zixuan~Chen,~
        Ruijie~Su,~
        Jiahao~Zhu,~
        Lingxiao~Yang,~\IEEEmembership{Member,~IEEE},~
        Jian-Huang~Lai,~\IEEEmembership{Senior Member,~IEEE},~
        and~Xiaohua~Xie,~\IEEEmembership{Member,~IEEE}
\thanks{Manuscript received XXX XX, XXXX; revised XXX XX, XXXX. This work was supported in part by the National Natural Science Foundation of China under Grant 62072482. (Corresponding author: Xiaohua Xie.)}
\thanks{The authors are with the School of Computer Science and Engineering, Sun Yat-sen University, Guangzhou 510006, China; and with the Guangdong Province Key Laboratory of Information Security Technology, Guangzhou 510006, China; and also with the Key Laboratory of Machine Intelligence and Advanced Computing, Ministry of Education, Guangzhou 510006, China. (e-mail: \{chenzx3, surj, zhujh59\}@mail2.sysu.edu.cn; \{yanglx9, stsljh, xiexiaoh6\}@mail.sysu.edu.cn)}
}

\markboth{IEEE TRANSACTIONS ON PATTERN ANALYSIS AND MACHINE INTELLIGENCE, VOL. XX, NO. XX, XXX. XXXX}%
{Shell \MakeLowercase{\textit{Chen et al.}}: VividDreamer: Towards High-Fidelity and Efficient Text-to-3D Generation}


\maketitle

\begin{abstract}
Text-to-3D generation aims to create 3D assets from text-to-image diffusion models.
However, existing methods face an inherent bottleneck in generation quality because the widely-used objectives such as Score Distillation Sampling (SDS) inappropriately omit U-Net jacobians for swift generation, leading to significant bias compared to the ``true'' gradient obtained by full denoising sampling.
This bias brings inconsistent updating direction, resulting in implausible 3D generation (\textit{e.g.,} color deviation, Janus problem, and semantically inconsistent details).
In this work, we propose \textit{Pose-dependent Consistency Distillation Sampling} (PCDS), a novel yet efficient objective for diffusion-based 3D generation tasks.
Specifically, \textit{PCDS} builds the \textit{pose-dependent consistency function} within diffusion trajectories, allowing to approximate true gradients through minimal sampling steps (1$\sim$3).
Compared to SDS, \textit{PCDS} can acquire a more accurate updating direction with the same sampling time (1 sampling step), while enabling few-step (2$\sim$3) sampling to trade compute for higher generation quality. 
For efficient generation, we propose a coarse-to-fine optimization strategy, which first utilizes 1-step \textit{PCDS} to create the basic structure of 3D objects, and then gradually increases \textit{PCDS} steps to generate fine-grained details.
Extensive experiments demonstrate that our approach outperforms the state-of-the-art in generation quality and training efficiency, conspicuously alleviating the implausible 3D generation issues caused by the deviated updating direction.
Moreover, it can be simply applied to many 3D generative applications to yield impressive 3D assets, please see \href{https://narcissusex.github.io/VividDreamer}{Project page}.
\end{abstract}


\section{Introduction}
\IEEEPARstart{T}{ext-to-3D} generation aims to create 3D objects with words, which used to sound like a tale from the \textit{Arabian Nights} as it relies on the organic combination between efficient 3D representation techniques and powerful generative models.
With the emergence of key technologies such as Neural Radiance Fields (NeRF) \cite{mildenhall2021nerf} and text-to-image diffusion models \cite{Rombach_2022_CVPR,saharia2022photorealistic}, numerous works \cite{poole2022dreamfusion,ProlificDreamer,lin2023magic3d,chen2023fantasia3d} quickly sprout up in recent years. 
These methods generate NeRF-based 3D representation from the 3D priors distilled from the pre-trained diffusion model, making it possible to create imaginative 3D assets in the real world.
To accelerate the training process, current methods \cite{tang2023dreamgaussian,yi2024gaussiandreamer,liang2024luciddreamer} introduce 3D Gaussian Splatting (3DGS) to replace NeRF in their frameworks, significantly reducing the optimization time from several hours to dozens of minutes.


However, existing text-to-3D generation models face an inherent bottleneck in generation quality.
Since the 3D priors are provided by the difference between rendered views and \textit{pseudo ground truth} (pseudoGTs, \textit{i.e.,} the denoised images), the acquisition of ``true'' gradients requires a full denoising sampling, leading to considerable sampling costs for each iteration.
To skip the full denoising sampling, Poole \etal and Liang \etal propose Score Distillation Sampling (SDS) and Interval Score Matching (ISM) objectives for swift generation, respectively.
Despite SDS and ISM significantly reducing the optimization time, it comes at the cost of generation quality as their estimated gradients often deviate from the true ones.
Specifically, to omit the terms of U-Net jacobians, SDS directly maps the noise to \textit{pseudoGTs} using 1-step DDPM sampling, which leads to significant bias and produces over-smoothing contents.
To address the over-smoothing issues, ISM first employs deterministic diffusing trajectories (\textit{i.e.,} DDIM inversion), and then simplifies the acquisition of true gradient into the interval scores at timestep $t$ through omitting the weighted sum of interval scores \textit{w.r.t} a series of timesteps.
While ISM performs better than SDS in generation quality, this omission changes the magnitude and direction of the true gradients, which brings inconsistent updating direction to 3D models.
Consequently, existing methods based on these objectives may lead to implausible and low-quality outcomes such as significant color deviation, frequently-occurred Janus problem, and semantically-inconsistent details (see \cref{sec:review} and \cref{fig:examples} in details).

\begin{figure*}[!t]
  \centering
  \includegraphics[width=0.9\textwidth]{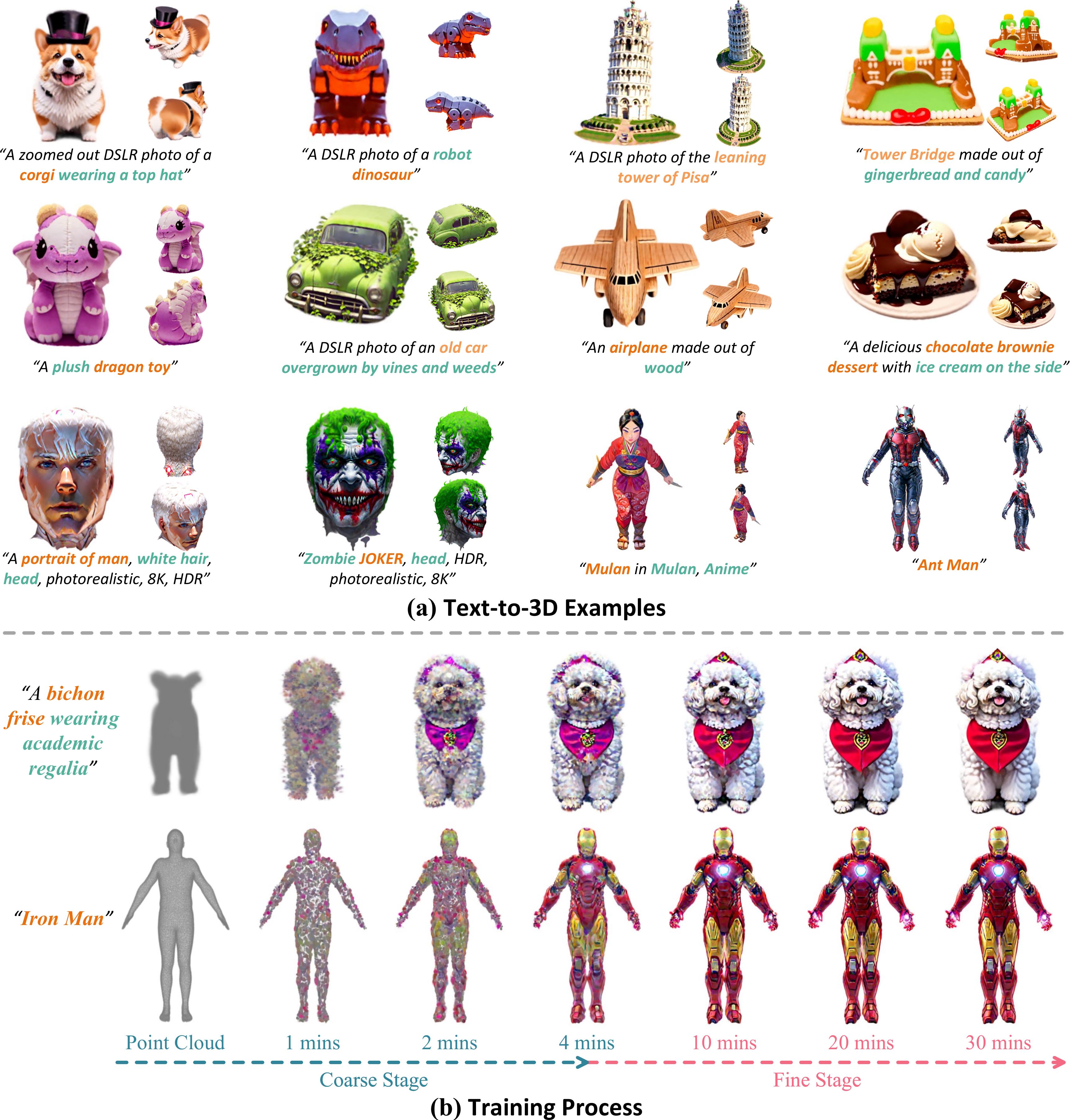}
  \caption{\textbf{Examples of text-to-3D asset creations with our framework (a).} We present an efficient text-to-3D generation framework -- \textbf{VividDreamer} that can distill semantically-consistent textures and high-fidelity structures from pretrained 2D diffusion models using a novel \textbf{Pose-dependent Consistency Distillation Sampling} objective in a \textbf{coarse-to-fine optimization} manner, allowing to yield high-fidelity 3D objects (rows 1 and 2) and 3D avatars (row 3) based on the given text prompts. Specifically, our \textbf{VividDreamer} achieves high training efficiency, which can create ready-to-use 3D assets within 10 minutes, while producing photorealistic 3D objects within 30 minutes \textbf{(b)}. More results can be found in \cref{fig:more} and our \href{https://narcissusex.github.io/VividDreamer}{Project page}.}
  \label{fig:first}
\end{figure*}


In this work, we propose \textit{VividDreamer}, an efficient framework that can effectively address the inherent bottleneck of text-to-3D generation.
Our key idea is to propose Pose-dependent Consistency Distillation Sampling (PCDS), a universal objective for diffusion-based 3D generation tasks.
Inspired by Consistency Models \cite{song2023consistency,luo2023latent}, our \textit{PCDS} builds the \textit{pose-dependent consistency function} within diffusion trajectories, enabling to skip the full denoising sampling and accurately estimate the gradients through 1$\sim$3 sampling steps.
Specifically, \textit{PCDS} first merges the pose-dependent scores based on Perp-Neg \cite{armandpour2023re}, and then maps the noise to \textit{pseudoGTs} using consistency function in a classifier-free guidance (CFG) manner \cite{ho2021classifier}.
Compared with existing objectives, \textit{PCDS} can acquire more accurate updating directions with the same sampling costs, while it further supports few-step (2$\sim$3) sampling to trade compute for better generation quality. 
For an efficient generation, we tailor a coarse-to-fine optimization strategy to achieve a better balance between training time and generation quality, which first utilizes 1-step \textit{PCDS} to create the basic structures, and then gradually increases \textit{PCDS} steps to generate fine-grained details.
We also introduce 3D Gaussian Splatting (3DGS) \cite{3dgs} to build our pipeline.
As a result, \textit{VividDreamer} achieves high training efficiency, enabling high-fidelity 3D object generation in a short time (see \cref{fig:first}).

Extensive experiments demonstrate that \textit{VividDreamer} favorably outperforms the state-of-the-art in generation quality, effectively alleviating the implausible 3D generation issues caused by deviated updating direction in existing methods (see \cref{fig:comp} in details).
\textit{VividDreamer} also achieves better training efficiency than the state-of-the-art, yielding higher generation quality with the same optimization time (see \cref{fig:ten_comp}).
\textit{Notably,} only 10 minutes of training on a single A100 GPU achieves superior generation quality to the existing 3DGS-based methods, being favored by most users in the user study (see \Cref{tab:us}).

Our framework has several other advantages: 
\textbf{i) High Consistency.} Thanks to the proposed \textit{PCDS} that can counteract the semantic deviation brought by intrinsic randomness, our \textit{VividDreamer} enables to yield consistent \textit{pseudoGTs} from various timesteps, significantly addressing the over-smoothing issues caused by SDS, producing more detailed results.
Moreover, unlike ISM which is only applicable to DDIM sampling, our \textit{PCDS} maintains such consistency in DDPM and DDIM sampling, 
\textbf{ii) High Adjustability.} Our \textit{PCDS} enables various-step sampling to meet different speed-quality demands. 
\textbf{iii) High Versatility.} Our \textit{PCDS} can be simply-applied into numerous 3D generative applications, creating high-quality 3D assets, please see \cref{fig:avatar,fig:editing} in details.

The main contributions are summarized as follows:
  
  
  
\begin{itemize}[nosep]
  \item We propose \textit{VividDreamer}, a novel framework for high-fidelity and efficient text-to-3D generation.
  
  \item We propose Pose-dependent Consistency Distillation Sampling (PCDS) to address the inherent bottleneck in generation quality, and also tailor a coarse-to-fine optimization strategy for efficient 3D generation.
  
  \item Experiments on various 3D generation tasks show that \textit{VividDreamer} favorably surpasses the state-of-the-art in generation quality and training efficiency.
\end{itemize}

The remainder of this paper is structured as follows. 
Related works and preliminaries are reviewed in \Cref{sec:rw,sec:pre}, respectively. \Cref{sec:methods} presents the analysis, motivation, and details of our approach. \Cref{sec:exp} demonstrates the experimental results and ablation study on text-to-3D generation. \Cref{sec:app} presents the results of other 3D generative applications. Conclusions and limitations are drawn in \Cref{sec:con}.

\section{Related Works}\label{sec:rw}
\subsection{Text-to-Image Diffusion Models}
The field of text-to-image generation has experienced rapid development in recent years, largely thanks to advances in diffusion models \cite{ho2020denoising,song2020score,song2023consistency} and CLIP \cite{radford2021learning}.
Being trained on a rich text-to-image dataset such as \cite{schuhmann2022laion}, large diffusion models \cite{nichol22a,Rombach_2022_CVPR,saharia2022photorealistic} are capable of generating impressive images consistent with a given text prompt conditioned on CLIP, becoming one of the core components for numerous currently-emerged generation and editing techniques.
Beyond basic text-conditioning, diffusion models can be further conditioned by various modalities.
One example is ControlNet \cite{zhang2023adding}, which can enhance the generation quality through the priors of depth or segmentation maps.
However, diffusion generation is inherently slow as it relies on an iterative sampling process for denoising.
This limits the development of downstream applications that require intensive queries, leading to massive training costs, especially in text-to-3D generation.

\subsection{Differentiable 3D Representations}
Creating a photorealistic 3D entity from discrete samples has been a long-standing research problem in the field of computer vision and graphics.
One of the commonly used solutions is Neural Radiance Fields (NeRF) \cite{mildenhall2021nerf}.
By utilizing standard volumetric rendering \cite{rendering} and alpha compositing techniques \cite{alpha}, NeRF builds a differentiable rendering pipeline, yielding impressive novel view synthesis from conventional photos.
A series of follow-up works extend NeRF \cite{mildenhall2021nerf} to the various applications, such as generative 3D modeling \cite{graf,poole2022dreamfusion}, 3D-editing \cite{nerf-editing,sine}, surface reconstruction \cite{wang2021neus}, and medical imaging \cite{chen2023cunerf}.
Other methods aim to enhance NeRF on training efficiency and reconstruction quality, \textit{e.g.,} few-view reconstruction \cite{pixnerf}, acceleration \cite{instngp}, and anti-aliasing \cite{mip-nerf}.
Recently, 3D Gaussian Splatting (3DGS) \cite{3dgs} has become a hot method in this field, achieving remarkable training efficiency.
In this paper, we introduce 3DGS to build the 3D rendering pipeline of our framework.

\subsection{Text-to-3D Generation Models}
With the breakthrough of key technologies in the fields of text-to-image generation and 3D representations, text-to-3D generation has become a reality.
As a pioneer, DreamFusion \cite{poole2022dreamfusion} first extracts the 3D priors given by pre-trained 2D text-to-image diffusion models to generate 3D NeRF objects.
Its core component is the Score Distillation Sampling (SDS), which facilitates 3D distillation by capturing the updating direction that conforms to text guidance, allowing for training a 3D model based on the 2D knowledge from diffusion models.
Motivated by SDS, numerous follow-up works \cite{ProlificDreamer,lin2023magic3d,chen2023fantasia3d,shi2023mvdream,richardson2023texture,zhu2023hifa,tang2023dreamgaussian,yi2024gaussiandreamer,liang2024luciddreamer} quickly sprout up to improve text-to-3D generation pipelines in various ways. 
Specifically, a group of methods \cite{lin2023magic3d,chen2023fantasia3d,tang2023dreamgaussian,yi2024gaussiandreamer} focus on better visual quality by modifying NeRF or introducing other efficient 3D representation techniques.
Other improvements aim at seeking an optimal distribution \cite{ProlificDreamer} or solving the \textit{Janus} problems \cite{shi2023mvdream}.
Albeit promising, SDS has shown over-smoothing effects in many papers \cite{poole2022dreamfusion,lin2023magic3d,richardson2023texture}, while its improvements like Interval Score Matching (ISM) \cite{liang2024luciddreamer} generates semantically-inconsistent details (\textit{e.g.,} color deviation), and others require a time-consuming sampling process \cite{ProlificDreamer,zhu2023hifa}.
Our work is intrinsically different in the sense that it provides a theoretical analysis of the inconsistency and low-quality generation caused by SDS and ISM.
We propose a universal approach Consistency Distillation Sampling (CDS), which can achieve superior generation quality with the same sampling costs as SDS and ISM, significantly addressing the above inherent bottleneck in text-to-3D generation.

\section{Preliminaries}\label{sec:pre}
\subsection{Diffusion Models}
\noindent\textbf{\textit{Denoising Diffusion Probabilistic Model (DDPM).}} DDPM \cite{ho2020denoising} assumes the inversion as a diffusion process according to a pre-defined schedule $\beta_t$ on timestep $t$ as:
\begin{equation}
  p(x_t|x_{t-1})=\mathcal{N}(x_t;\sqrt{1-\beta_t}x_{t-1},\beta_t\mathit{I}).
  \label{eq:ddpm}
\end{equation}
The posterior is modeled by a neural network $\phi$ as:
\begin{equation}
  p_\phi(x_{t-1}|x_t)=\mathcal{N}(x_{t-1};\sqrt{\overline{\alpha}_{t-1}}\mu_\phi(x_t),(1-\overline{\alpha}_{t-1})\Sigma_\phi(x_t)),
  \label{eq:ddpm_pre}
\end{equation}
where $\overline{\alpha}_t:=(\prod^t_1(1-\beta_t))$, while $\mu_\phi$ and $\Sigma_\phi(x_t)$ denote the predicted mean and variance of $x_t$, respectively. 

\noindent\textbf{\textit{Denoising Diffusion Implicit Model (DDIM).}} Given an inversion trajectory $\{0, k,2k,\cdots,t-k,t\}$, DDIM \cite{ddim} predicts invertible noisy latents as a deterministic diffusion trajectories:
\begin{equation}
  x_t = \sqrt{\overline{\alpha}_t}\hat{x}^{t-k}_0+\sqrt{1-\overline{\alpha}_t}\epsilon_\phi(x_{t-k},t-k,\emptyset),
  \label{eq:ddim}
\end{equation}
where $\hat{x}^{t-k}_0=\frac{1}{\sqrt{\overline{\alpha}_s}}x_{t-k}-\sqrt{1-\overline{\alpha}_t}\epsilon_\phi(x_{t-k},t-k,\emptyset)$ denotes the \textit{pseudoGTs} directly estimated from $x_{t-k}$ using 1-step sampling.
Thus, any point within that deterministic diffusion trajectory can be non-randomly estimated iteratively. 

\subsection{3D Gaussian Splatting} 
3D Gaussian Splatting (3DGS) \cite{3dgs} is a recent groundbreaking method for novel view synthesis.
Unlike NeRF \cite{mildenhall2021nerf} and its variants that render images based on volume rendering \cite{rendering}, 3DGS renders images through splatting \cite{splatting}, allowing extremely fast optimization and rendering speed.
Specifically, 3DGS represents the scene through a set of 3D Gaussians.
These 3D Gaussians can be defined with its center position $\mu\in\mathbb{R}^3$, covariance $\varSigma\in\mathbb{R}^7$, color $c\in\mathbb{R}^3$, and opacity $\alpha\in\mathbb{R}^1$, which can be queried as:
\begin{equation}
  \mathcal{G}(p)=\exp(-\frac{1}{2}(p)^T\varSigma^{-1}(p)),
\end{equation}
where $p$ denotes the distance between $\mu$ and the query point.
For computing the color of each pixel, 3DGS uses neural point-based rendering \cite{splatting}, \textit{i.e.,} simulating a ray $r$ casting from the center of the camera to the end based on the corresponding camera pose $\mathbf{c}$, which can be formulated as:
\begin{equation}
  C(r) = \sum_{i\in\mathcal{N}}c_i\sigma_i\prod_{j=i}^{i-1}(1-\sigma_j),\quad \sigma_i=\alpha_i\mathcal{G}(p_i), \label{eq:3dgs_rendering}
\end{equation}
where $\mathcal{N}$ indicates the samples on the ray $r$, while $\sigma_i$, $c_i$, $\alpha_i$ and $p_i$ denote the density, color, opacity, and distance of the $i$-th 3D Gaussians, respectively.
Since the rendering process is differentiable, 3D Gaussians can be optimized using gradient descent.
After optimization, 3DGS projects 3D Gaussians to 2D for rendering, yielding photorealistic novel views.
Given a viewing transformation $W$, the covariance matrix $\varSigma'$ in camera coordinates can be acquired as:
\begin{equation}
  \varSigma'=JW\varSigma W^TJ^T,
\end{equation}
where $J$ is the Jacobian of the affine approximation of the projective transformation.

\begin{figure*}[!t]
  \centering
  \includegraphics[width=0.8\textwidth]{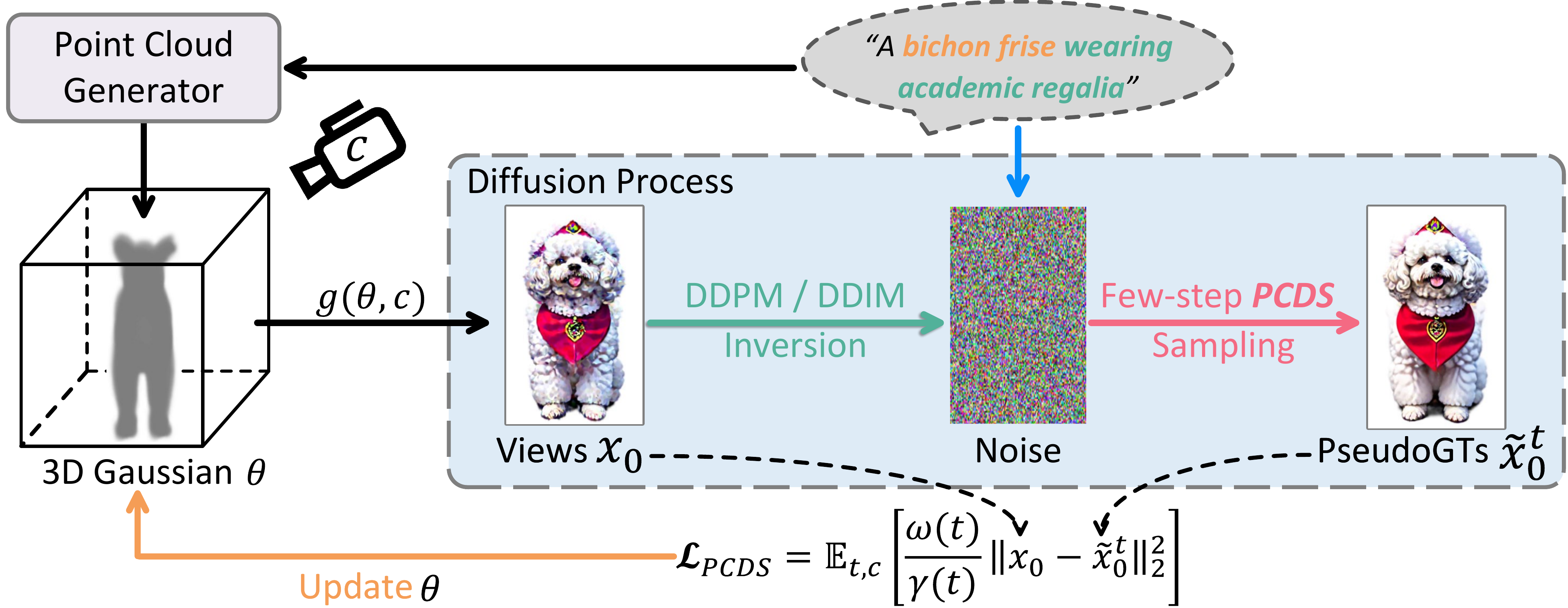}
  \caption{
    An overview of \textit{VividDreamer}. 
    We employ 3D Gaussian Splatting (3DGS) \cite{3dgs} as 3D representation, and initialize it using the pre-trained Point-E \cite{point-e} with given text prompts. 
    In training, given a camera pose $c$, we render the corresponding view $x_0=g(\theta,c)$ by the rendering pipeline of 3DGS, and disturb it to 2D diffusion models using DDPM/DDIM inversion. 
    Then, we employ the proposed Pose-dependent Consistency Distillation Sampling (PCDS) to map noise $x_t$ to the \textit{pseudoGTs} $\tilde{x}^t_0$ (\textit{i.e.,} the denoised images) through few-step (1$\sim$3) sampling.
    Finally, we calculate the Mean Square Error (MSE) loss $\mathcal{L}_{PCDS}$ between the rendered views $x_0$ and \textit{pseudoGTs} $\tilde{x}^t_0$, and update the parameter of 3D Gaussians $\theta$ by the gradients $\bigtriangledown_\theta\mathcal{L}_{PCDS}$ in \cref{eq:cds_consistency}.
  }
  \label{fig:examples}
\end{figure*}

\section{Methods}\label{sec:methods}
In this section, we first analyze the limitations of Score Distillation Sampling (SDS) and Interval Score Matching (ISM) in text-to-3D generation.
Subsequently, inspired by Consistency Models \cite{song2023consistency,luo2023latent} we propose Pose-dependent Consistency Distillation Sampling, a novel yet efficient objective that can acquire accurate gradients with minimal step sampling (1$\sim$3), significantly addressing the bottleneck of previous works in generation quality.
Finally, we propose a two-stage framework for efficient text-to-3D generation, allowing high-fidelity 3D asset creation in a short training time.
Figure \ref{fig:examples} depicts the overall framework of our \textit{VividDreamer}, and the subsequent techniques are described in the following subsections. 

\subsection{Review of SDS and ISM.}\label{sec:review}
Given a 3D representation with learnable parameter $\theta$, the differentiable rendering $x_0=g(\theta, \mathbf{c})$ denotes to render 2D images $x_0$ based on the corresponding camera poses $\mathbf{c}$.
As shown in \cref{fig:difference} \textbf{(a)}, the true gradients can be calculated as:
\begin{equation}
\bigtriangledown_\theta\mathcal{L}_{true}(\theta) = \mathbb{E}_{t,\mathbf{c}}\left[\frac{\omega(t)}{\gamma(t)}(x_0-\tilde{x}^t_0)\frac{\partial g(\theta,\mathbf{c})}{\partial \theta}\right],
\label{eq:true}
\end{equation}
where $\omega(t)$ and $\gamma(t)=\frac{\sqrt{1-\bar{\alpha}_t}}{\sqrt{\bar{\alpha}_t}}$ are the weight \textit{w.r.t} different timesteps.
$\tilde{x}^t_0$ denotes the \textit{pseudoGTs} obtained by full denoising sampling from the noise $x_t$ to the timestep $0$.
Since the acquisition of $\tilde{x}^t_0$ takes considerable sampling costs, SDS and ISM objectives aim to skip such a time-consuming process, which can be detailedly analyzed as follows:

\noindent\textbf{\textit{Score Distillation Sampling (SDS).}} 
As shown in \cref{fig:examples} \textbf{(b)}, SDS objective can be formulated as:
\begin{equation}
  \bigtriangledown_\theta\mathcal{L}_{SDS}(\theta) = \mathbb{E}_{t,\mathbf{c}}\left[\frac{\omega(t)}{\gamma(t)}(x_0-\hat{x}^t_0)\frac{\partial g(\theta,\mathbf{c})}{\partial \theta}\right],
\end{equation}
where $\hat{x}^t_0$ denotes the \textit{pseudoGTs} directly estimated from the noise $x_t$ using 1-step DDPM sampling based on \cref{eq:ddpm_pre}.
Due to the intrinsic randomness brought by DDPM, SDS faces two inherent limitations: \textit{i)} inconsistent $\hat{x}^t_0$ across different timesteps $t$; and \textit{ii)} inaccurate 1-step estimation for all $t$, leading to blurred and sketchy results.

\noindent\textbf{\textit{Interval Score Matching (ISM).}} Unlike SDS, Liang \etal non-randomly estimate $x_t$ using DDIM inversion in \cref{eq:ddim}, and thus \cref{eq:true} can be rewritten as:
\begin{equation}
  \bigtriangledown_\theta \mathcal{L}_{true}(\theta)\!\triangleq\!\mathbb{E}_{t,\mathbf{c}}[(\omega(t)[\underbrace{\epsilon_\phi(x_t,\!t,\!y)\!-\!\epsilon_\phi(x_s,\!s,\!\emptyset)}_{interval\ score}]+\eta_t\!)\!\frac{\partial g(\theta,\!\mathbf{c})}{\partial \theta}],
  \label{eq:true_grads}
\end{equation}
where $s=t-k$, and the bias term $\eta_t$ is a weighted sum of interval scores \textit{w.r.t} a series of timesteps.
To skip full denoising sampling, ISM objective $\bigtriangledown_\theta\mathcal{L}_{ISM}$ directly omits the bias term $\eta_t$ and only maintains the interval score at timestep $t$ as:
\begin{equation}
  \bigtriangledown_\theta \mathcal{L}_{ISM}(\theta)=\mathbb{E}_{t,\mathbf{c}}[(\omega(t)[\epsilon_\phi(x_t,t,y)-\epsilon_\phi(x_s,s,\emptyset)])\frac{\partial g(\theta, \mathbf{c})}{\partial \theta}],
  \label{eq:ISM}
\end{equation}
Since such omission changes the magnitude and direction of the true gradients $\bigtriangledown_\theta \mathcal{L}_{true}$, it leads to implausible outcomes, \textit{e.g.}, color deviation, Janus problems, and semantically-inconsistent details in 3D models (see \cref{fig:comp,fig:loss_comp}).

\subsection{Pose-dependent Consistency Distillation Sampling}
\begin{figure}[!t]
  \centering
  \includegraphics[width=0.48\textwidth]{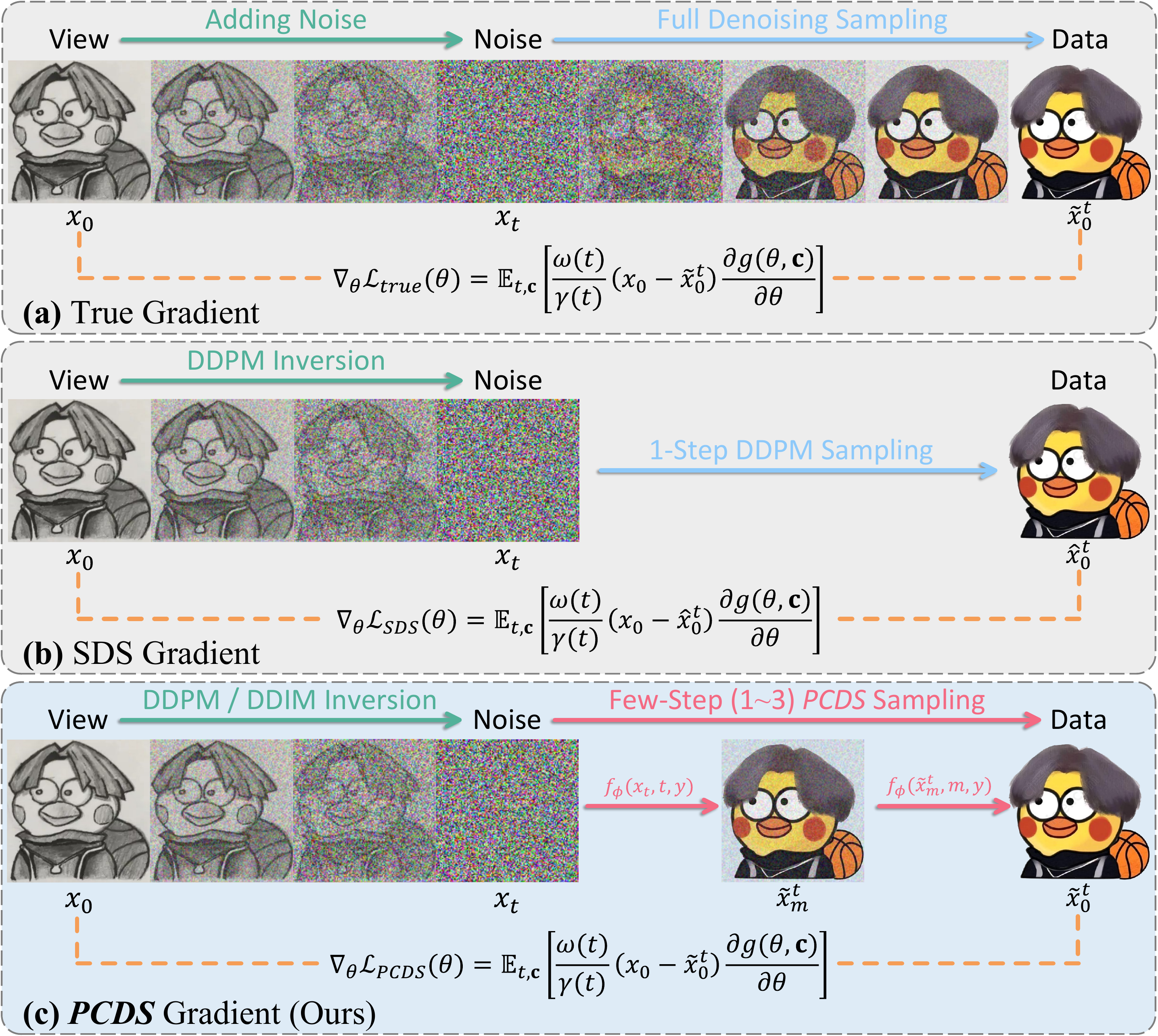}
  \caption{
  Examples of different objectives. 
  Visually, the acquisition of ``true'' gradient \textbf{(a)} is time-consuming work, requiring the full denoising sampling in each iteration.
  To skip such a lengthy process, Score Distillation Sampling (SDS) \cite{poole2022dreamfusion} \textbf{(b)} directly maps the noise to data \textit{i.e., pseudoGTs} using 1-step DDPM sampling, but SDS struggles to acquire accurate gradients due to the intrinsic randomness brought by DDPM.
  On the contrary, our \textit{PCDS} builds the \textit{pose-dependent consistency function} $f_\phi$ from any timestep $t$ to the origin $0$ within diffusion trajectories, allowing to generate accurate \textit{pseudoGTs} and acquire precise gradients via minimal sampling steps (1$\sim$3).
  }
  \label{fig:difference}
\end{figure}

To mitigate the above problems exposed by SDS and ISM, we propose a novel sampling strategy dubbed Pose-dependent Consistency Distillation Sampling (PCDS) for text-to-3D generation. 
As analyzed in \cref{sec:review}, the core problem of ISM is how to efficiently obtain consistent estimation $\tilde{x}_0^t$ from any $x_t$. 
To this end, we only need to build a \textit{consistency function} that can output consistent estimation $\tilde{x}_0^t$ from any $x_t$ on a certain trajectory with few-step (1 $\sim$ 3) sampling (see \cref{fig:difference} \textbf{(c)}). 
Inspired by \cite{luo2023latent}, this function can be formally defined as $f_\phi: (x_t, t, y)\to \tilde{x}^t_0$, which can be further parameterized by:
\begin{equation}
  f_\phi=f_{in}(t)x_0+f_{out}(t)\mathcal{C}_\phi(x_t,t,y),
\end{equation}
where $f_{in}$ and $f_{out}$ are differentiable functions with $f_{in}(0)=1$ and $f_{out}(0)=0$, and $\mathcal{C}_\phi$ is a neural network that outputs a tensor with the size of $x_t$. 
Concretely, we set $\mathcal{C}_\phi(x_t, t, y)$ as:
\begin{equation}
  \mathcal{C}_\phi(x_t, t, y)=\frac{x_t}{\sqrt{\bar{\alpha}_t}}-\gamma(t)\hat{\epsilon}_\phi(x_t,t,y),\ t>0,
\end{equation}
which can be regarded as 1-step DDIM sampling \cite{ddim}. 
Since $\tilde{x}^t_0$ can be accurately estimated from $x_t$ through few-step \textit{PCDS} sampling, we can obtain an accurate and fast approximation of the true gradient $\bigtriangledown_\theta \mathcal{L}_{true}$ as:
\begin{equation} 
     \bigtriangledown_\theta \mathcal{L}_{PCDS}(\theta)\triangleq\mathbb{E}_{t,\mathbf{c}}\left[\frac{\omega(t)}{\gamma(t)}(x_0-\tilde{x}^t_0) \frac{\partial g(\theta,\mathbf{c})}{\partial \theta}\right].
    \label{eq:cds_consistency}
\end{equation} 

Though classifier-free guidance (CFG) \cite{ho2021classifier} has been widely used to construct $\hat{\epsilon}_\phi(x_t,t,y)$ in text-to-image generation, directly applying it to text-to-3D generation may bring about the notorious Janus problem, \textit{e.g.,} yielding multi-faced objects.
To further alleviate this problem, we incorporate pose-dependent prompts following Perp-Neg \cite{armandpour2023re} into our \textit{PCDS}.
Specifically, given a camera pose $\mathbf{c}$, we generate its pose-dependent text embedding with the interpolation technique proposed in Perp-Neg. 
Then, we construct $\hat{\epsilon}_\phi(x_t,t,y)$ as follows:
\begin{equation}
  \hat{\epsilon}_\phi(x_t,t,y)=\epsilon_\phi(x_t,t)+w_g\left[\epsilon^{pos}_\phi-\sum_{i=1}^{N_{neg}}w_{c}^{(i)}\epsilon^{neg^{(i)}\perp}_\phi\right],
  \label{eq:eps_perp_neg}
\end{equation}
where $N_{neg}$ denotes the number of negative prompts embeddings \textit{w.r.t} camera pose $c$, while $w_g$ and $w_c^{(i)}$ indicate the weighting coefficients of positive and negative prompts embeddings, respectively.
And $\epsilon^{pos}$ and $\epsilon^{neg^{(i)}}_\phi$ are constructed from the unconditional term $\epsilon_\phi(x_t,t)$ as:
\begin{equation}
  \begin{split}
  \epsilon^{pos}_\phi=\epsilon_\phi(x_t,t,y_c^{pos})-\epsilon_\phi(x_t,t),\\
  \epsilon^{neg^{(i)}}_\phi=\epsilon_\phi(x_t,t,y_c^{neg^{(i)}})-\epsilon_\phi(x_t,t),
  \end{split}
\end{equation}
where $y_c^{pos}$ and $y_c^{neg^{(i)}}$ denote the positive and negative prompt embeddings \textit{w.r.t} camera pose $c$, respectively.
Thus, the perpendicular gradient $\epsilon^{neg^{(i)}\perp}_\phi$ of $\epsilon^{pos}_\phi$ on $\epsilon^{neg^{(i)}}$ can be estimated as:
\begin{equation}
  \epsilon^{neg^{(i)}\perp}_\phi=\left(\epsilon^{neg^{(i)}}_\phi-\frac{\left<\epsilon^{pos}_\phi,\epsilon^{neg^{(i)}}_\phi\right>}{\|\epsilon^{pos}_\phi\|^2}\epsilon^{pos}_\phi\right).
\end{equation}

In practice, our \textit{PCDS} first obtain the pose-dependent scores by \cref{eq:eps_perp_neg}, and then iteratively estimates the \textit{pseudoGTs} in a CFG manner, which achieves robust sampling against Janus problem, enabling high-fidelity 3D object creation.


\subsection{Coarse-to-Fine Optimization}
As discussed, \textit{PCDS} enables directly mapping the noise to \textit{pseudoGTs} through 1-step sampling, while also allowing few-step (2$\sim$3) sampling to trade compute for higher generation quality.
For efficient text-to-3D generation, we tailor a two-stage optimization strategy, which can generate 3D assets from coarse to fine, achieving high training efficiency.
For clarity, we also provide pseudo codes in \Cref{alg:ctf}.

\noindent\textit{In} \textbf{\textit{Coarse Stage}}, we aim to rapidly create an initial 3D structure from the given point cloud. 
For fast convergence, we first map the rendered views $x_0$ to noise $x_t$ using DDPM inversion in \cref{eq:ddpm}, and then directly predict the \textit{pseudoGTs} via a 1-step \textit{PCDS} sampling.
Thanks to the proposed \textit{PCDS}, only 1-step sampling can acquire precise gradients for 3D object initialization, yielding better generation quality than SDS (see \cref{fig:loss_comp}).
The coarse stage only takes 4 minutes of training to create a nice structure from initial point clouds as shown in \cref{fig:first}, effectively alleviating bizarre shapes caused by intrinsic randomness in the early training.

\noindent\textit{In} \textbf{\textit{Fine Stage}}, we aim to refine the 3D structures and generate fine-grained details.
Specifically, to acquire more precise gradients, we utilize the DDIM inversion \cref{eq:ddim} instead of the DDPM inversion to non-randomly map the rendered views $x_0$ to noise $x_t$.
We also gradually increase the sampling steps of \textit{PCDS} from 1 to 3 with the increasing optimization steps, allowing the model to learn more fine-grained details from the diffusion-based 3D priors.
As a result, \textit{VividDreamer} can create ready-to-use 3D assets within 10 minutes, and producing photorealistic objects within 30 minutes (see \cref{fig:first}).

\subsection{Advanced Generation Pipeline}
We also explore some factors that would affect the generation quality and propose an advanced pipeline with our \textit{PCDS}. 
Specifically, we propose an efficient 3D representation technique and utilize ControlNet \cite{zhang2023adding} for enhancement.

\subsubsection{Efficient Initialization}
Existing 3DGS-based methods usually adopt initial point cloud provided by text-to-point generative models \cite{point-e,jun2023shap} for fast convergence.
However, these text-to-point generative models cannot understand complex text prompts, usually yielding strange and irregular point clouds.
Thus, the acquisition of the initial point cloud is a labor-consuming task because it requires manual conversion of text prompts into simple words, \textit{e.g.,} converting ``A hi-poly model of a yellow supercar'' into ``car''.
To address this issue, we utilize Chain-of-Thought (COT) \cite{wei2022chain} to teach the ChatGPT \cite{ChatGPT} how to extract simple prototypes from a complex text description.
After training, ChatGPT can simplify a complex text description into one or two simple words.
These words can be directly fed into the text-to-point generative models, obtaining point clouds with similar semantics, allowing to liberate researchers from heavy labor in manual conversions.

\subsubsection{ControlNet-based Enhancement} \label{sec:agp}
Janus problem is a great challenge in text-to-3D generation tasks. 
Though our \textit{PCDS} significantly alleviates the Janus problem caused by inaccurate gradients, it still cannot completely solve this problem because the text-to-image diffusion models do not truly understand the 3D objects in the real world.
To achieve better 3D consistency, we optionally introduce ControlNet \cite{zhang2023adding} to guide diffusion models, where the ControlNet priors are based on the depth maps related to the camera poses.
This can successfully alleviate the Janus problem that appears in some specific text prompts, enabling better robustness against various situations. 

\begin{algorithm}[!t]
  \caption{Coarse-to-Fine Optimization}
  \label{alg:ctf}
  \SetAlgoLined
  \SetKwInOut{Input}{Input}
  \SetKwInOut{Output}{Output}
  \Input{Consistency model $\mathcal{C}_\phi$, Coarse-stage iteration number $N_c$, Fine-stage iteration number $N_f$, DDIM inversion stepsize $\delta_t$, and Target prompt embedding $y$.}
  \BlankLine
  \For{$i \in \left\{0, 1, \cdots, N_c+N_f-1\right\}$}{
    \If{$i<N_c$}{
      \textit{Sample:} $x_0=g(\theta,c)$ and $t\sim \mathcal{U}[600, 700]$;\\
      \tcc{1-step DDPM inversion}
      $x_t = \sqrt{\bar{\alpha}_t}x_0+\sqrt{1-\bar{\alpha}_t}\epsilon$, $\epsilon\sim\mathcal{N}(\mathbf{0},\mathbf{I})$;\\
      $\tilde{x}_0^t=\mathcal{C}_\phi(x_t,t,y)$;\tcp*[h]{\footnotesize 1-step \textit{PCDS} sampling}\\
    }
    \Else{
      \textit{Sample:} $x_0=g(\theta,c)$ and $t\sim \mathcal{U}[300, 500]$;\\
      \textit{Initialize:} $x_t=x_0$;\\
      \textit{Let} $N_{inv} = \lceil t / \delta_t\rceil$;\\
      \tcc{DDIM inversion in \cref{eq:ddim}}
      \For{$j \in \left\{0, 1,\cdots, N_{inv} - 1\right\}$}{
        \textit{Let} $t_f\!=\!\min(\lfloor(j+1)\delta_t\rfloor,t)$ and $t_n\!=\!\lfloor j\delta_t\rfloor$;\\
        $\hat{x}^{t_n}_0=\frac{1}{\sqrt{\bar{\alpha}_{t_n}}}(x_t-\sqrt{1-\bar{\alpha}_{t_n}}\epsilon_\phi(x_t, t_n, \emptyset))$;\\
        $x_t \!\!\leftarrow\!\! \sqrt{\bar{\alpha}_{t_f}}\hat{x}^{t_n}_0\!\!+\!\!\sqrt{1\!-\!\bar{\alpha}_{t_f}}\epsilon_\phi(x_t,\! t_n,\! \emptyset)$;\\
      }
      \textit{Initialize:} $\tilde{x}_0^t=\mathcal{C}_\phi(x_t,t,y)$;\\
      \textit{Let} $N_p\in[1, 3]$ denote \textit{PCDS} steps;\\
      \tcc{Gradually increasing $N_p$}
      \If{$N_p > 1$}{
        \textit{Let} $\delta_p=t/N_p$;\\
        \For{$k\in\{N_p - 1, N_p - 2, \cdots, 1\}$}{
          \textit{Let} $t_f=\lfloor k\delta_p\rfloor$ and $t_n=\lfloor(k+1)\delta_p\rfloor$;\\
          $x_{t_f} \!\!=\!\! \sqrt{\bar{\alpha}_{t_f}}\tilde{x}_0^t+\!\sqrt{1-\bar{\alpha}_{t_f}}\epsilon_\phi(x_{t_n},t_n,y)$;\\
          $\tilde{x}_0^t\leftarrow\mathcal{C}_\phi(x_{t_f},t_f,y)$;\\
        }
      }
    }      
      Update $\theta$ according to \cref{eq:cds_consistency};
  }
\end{algorithm}

\begin{figure*}[!t]
  \centering
  \includegraphics[width=0.95\textwidth]{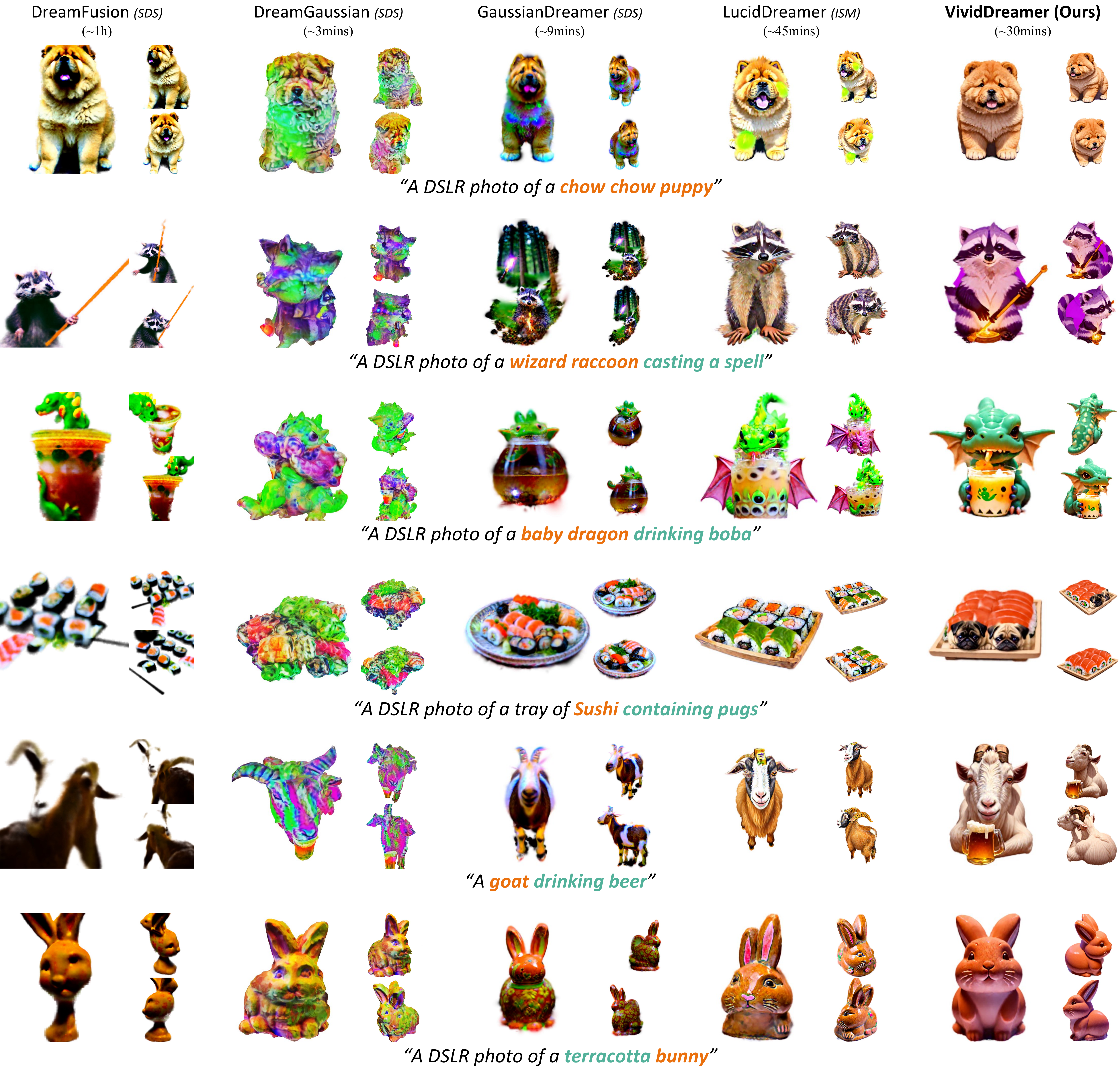}
  \caption{
    Visual comparisons between our framework and 4 state-of-the-art methods for text-to-3D generation. Experimental results show that our approach is capable of creating high-fidelity 3D assets that maintain consistent semantics with the given text prompts, significantly alleviating the color deviation, Janus problem, and semantically inconsistent details caused by inaccurate gradient estimation. The training time is evaluated on a single A100 GPU.
    }
  \label{fig:comp}
\end{figure*}

\begin{figure*}[!t]
  \centering
  \includegraphics[width=0.9\textwidth]{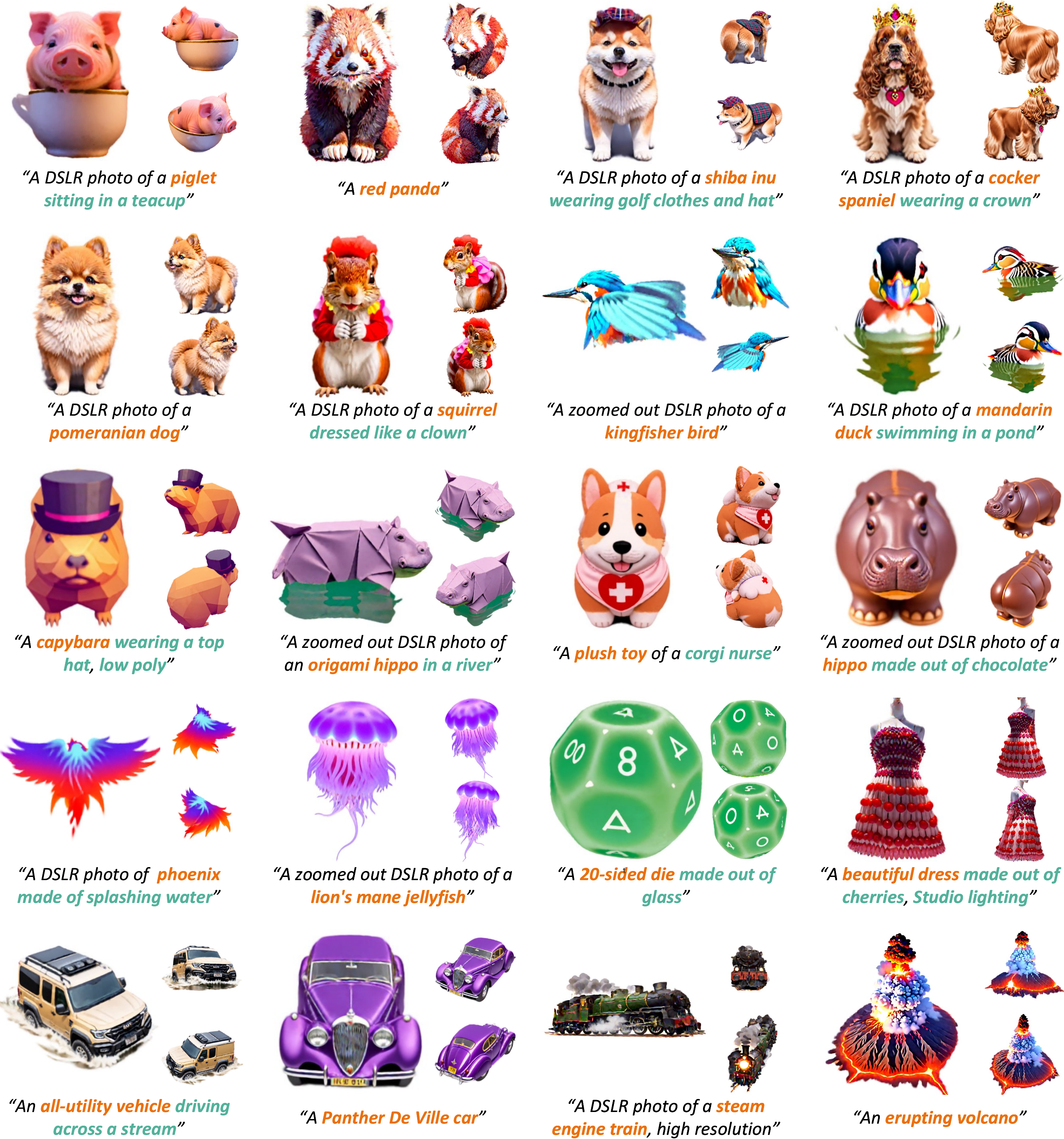}
  \caption{
    Visual results generated by our \textit{VividDreamer} framework with 30 minutes of training on a single A100 GPU. As shown, our approach creates high-fidelity 3D assets based on various text prompts. More visual results can be found in our \href{https://narcissusex.github.io/VividDreamer}{Project page}.
    }
  \label{fig:more}
\end{figure*}

\section{Experiments}\label{sec:exp}
In this section, we conduct extensive experiments and in-depth analysis to demonstrate the superiority of our \textit{VividDreamer} in text-to-3D generation.

\subsection{Implementation Details}
In our framework, we adopt 3D Gaussian Splatting (3DGS) \cite{3dgs} as 3D representation and Stable Diffusion \cite{Rombach_2022_CVPR} fintuned by Latent Consistency Model \cite{luo2023latent} as diffusion-based 3D priors.
We utilize the cloud point provided by the pre-trained Point-E \cite{point-e} for initialization, and feed the camera-dependent prompts suggested in Perp-Neg \cite{armandpour2023re} into diffusion models.
To further alleviate the Janus problem using ControlNet \cite{zhang2023adding}, we first render the depth maps related to the camera poses using 3DGS, and then feed these rendered depth maps into ControlNet to guide the generation of diffusion models.
The batch size is set to 4, and the iteration numbers of coarse stage $N_c$ and fine stage $N_f$ are set to 500 and 2500, respectively.
The DDIM inversion stepsize $\delta_t$ is fixed as 200, and the iteration numbers of 1-, 2-, and 3-step \textit{PCDS} sampling are set to 1000, 800, and 700, respectively.
\textit{Note}, we can acquire ready-to-use 3D assets within 10 minutes (nearly 1500 iterations), and obtaining high-fidelity objects within 30 minutes (3000 iterations).

\subsection{Comparison Results}
We primarily compare with four baselines, including DreamFusion \cite{poole2022dreamfusion}, DreamGaussian \cite{yi2024gaussiandreamer}, GaussianDreamer \cite{yi2024gaussiandreamer} and LucidDreamer \cite{liang2024luciddreamer}, where DreamFusion is re-implemented by \cite{stable-dreamfusion} using NGP \cite{instngp} and Stable Diffusion \cite{Rombach_2022_CVPR}, and the rest methods are built by 3DGS and Stable Diffusion.

\subsubsection{Qualitative Comparisons}
We provide qualitative comparisons on text-to-3D generation in \cref{fig:comp}. 
As shown, our \textit{VividDreamer} outperforms the competitors especially \textit{w.r.t} high-quality visual appearance, significantly alleviating the color deviation caused by inaccurate gradients.
We also found that directly using Perp-Neg \cite{armandpour2023re} can only alleviate the Janus problem that appears in some text prompts.
For example, though GaussianDreamer \cite{yi2024gaussiandreamer} and LucidDreamer \cite{liang2024luciddreamer} incorporate Perp-Neg into their 3D generation pipeline, they still produce multi-faced and 3D-inconsistent objects.
Thanks to \textit{PCDS} that neatly combines Perp-Neg into the built-in consistency function, \textit{VividDreamer} can effectively alleviate the Janus problem brought by inaccurate gradients and the intrinsic limitation of text-to-image diffusion models, yielding lifelike 3D assets with high-fidelity geometric structures.
Moreover, \textit{VividDreamer} generates semantically consistent results, faithfully restoring the details overlooked by the compared methods.
We also provide more visual results in \cref{fig:more}, and more visual comparisons can be found in our \href{https://narcissusex.github.io/VividDreamer}{Project page}.

\begin{figure}[!t]
  \centering
  \includegraphics[width=0.48\textwidth]{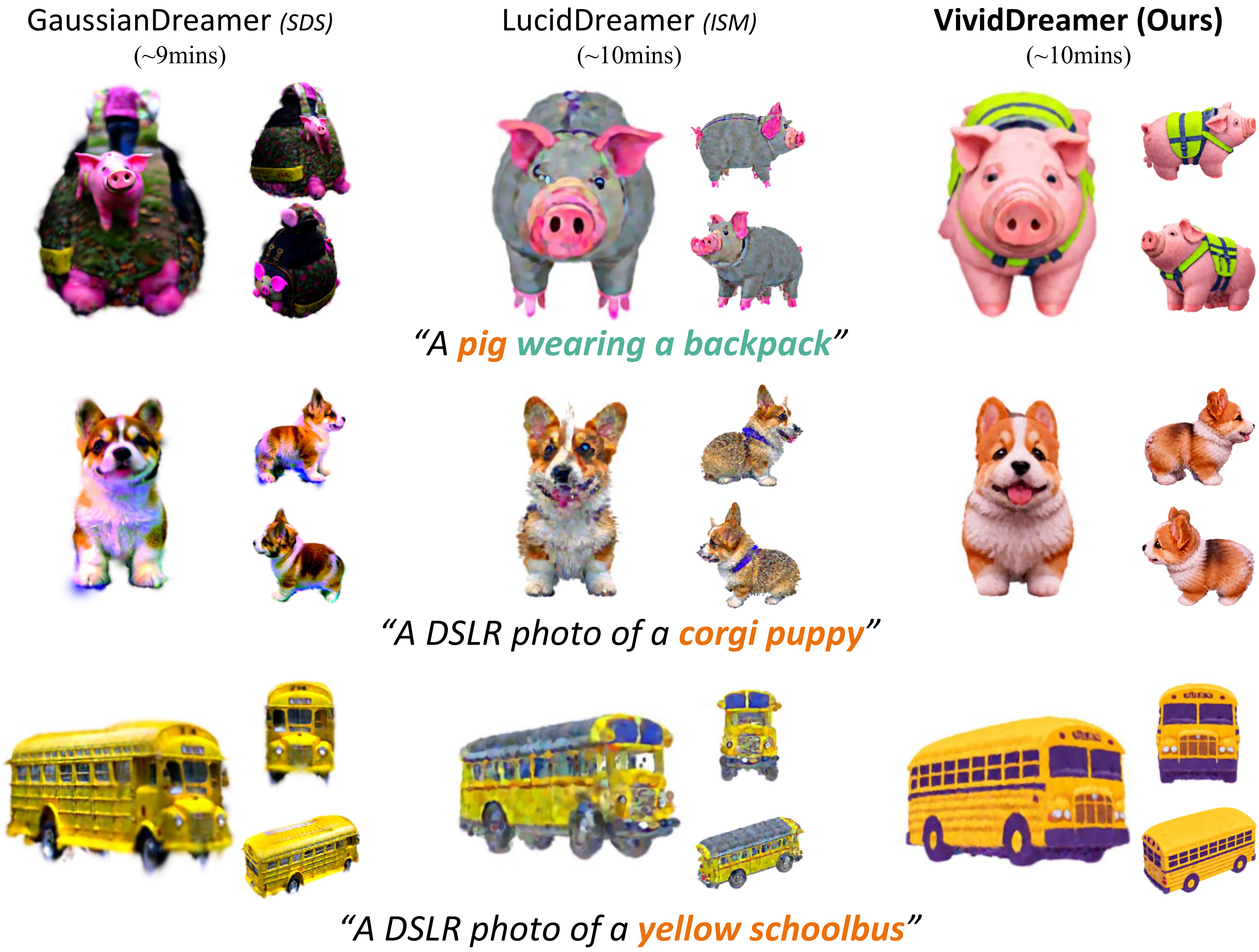}
  \caption{
    Visual comparisons between our \textit{VividDreamer} and 2 state-of-the-art methods: LucidDreamer \cite{liang2024luciddreamer}, and GaussianDreamer \cite{yi2024gaussiandreamer} for text-to-3D generation, where the results are generated by $\sim$10 minutes of training on a single A100 GPU.
    Visually, our framework enables high-fidelity 3D object creations within a such short time, presenting a noteworthy training efficiency.
  }
  \label{fig:ten_comp}
\end{figure}

\begin{figure}[!t]
  \centering
  \includegraphics[width=0.48\textwidth]{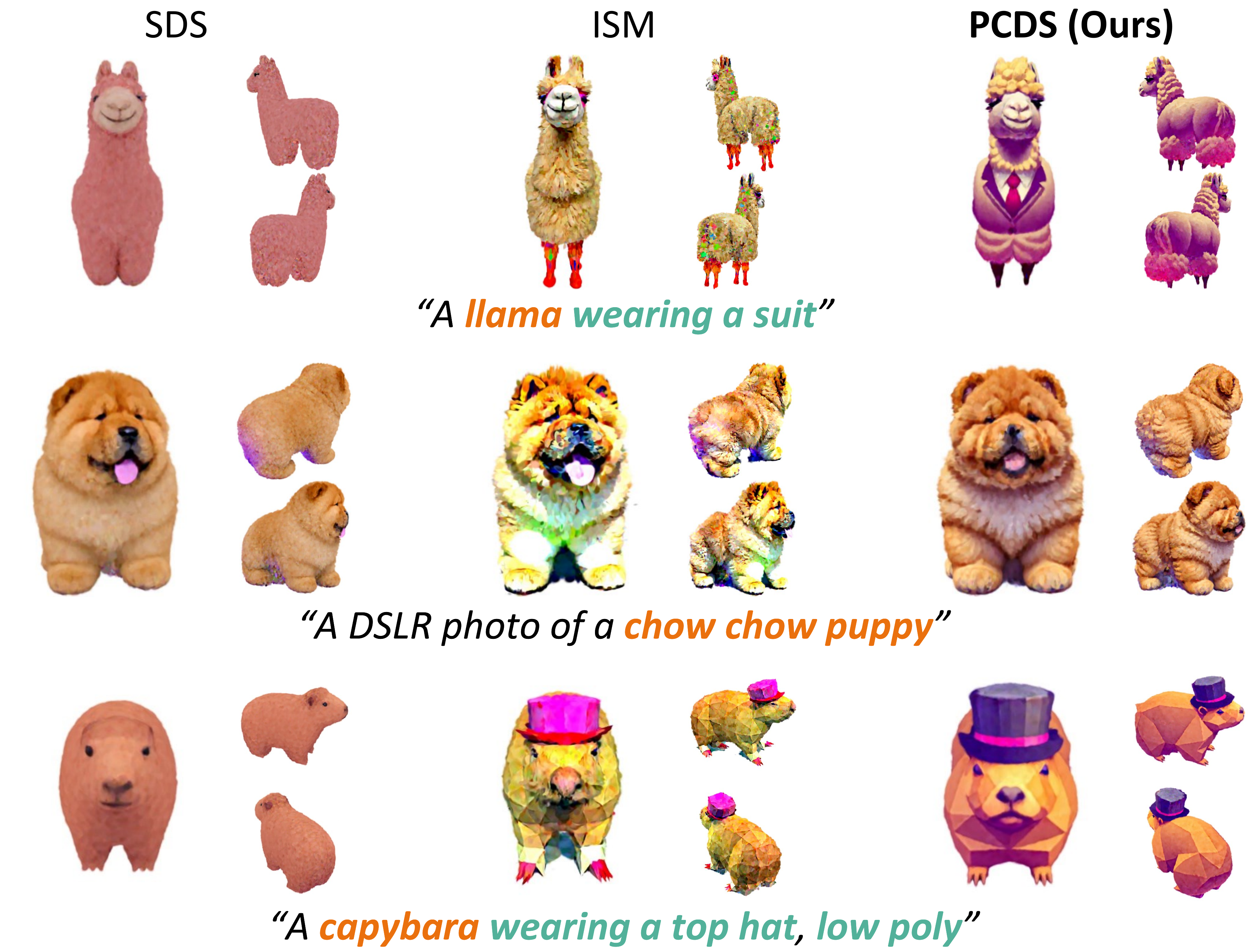}
  \caption{
    Visual comparisons between our Pose-dependent Consistency Distillation Sampling \textit{PCDS} and 2 existing objectives: Score Distillation Sampling (SDS) \cite{poole2022dreamfusion} and Interval Score Mathcing (ISM) \cite{liang2024luciddreamer}. 
    The results are generated by incorporating these objectives into our framework with a 30-minute training without any modifications and extra processing.
  }
  \label{fig:loss_comp}
\end{figure}

To better exhibit our training efficiency, we further compare the results against baselines under a 10-minute training.
As shown in \cref{fig:ten_comp}, LucidDreamer \cite{liang2024luciddreamer} falls short in good convergence within 10 minutes of training, yielding sub-optimal results.
While GaussianDreamer \cite{yi2024gaussiandreamer} can be well-converged in 10 minutes, it is limited by the inaccurate gradient estimation from SDS, producing low-quality 3D objects.
On the contrary, \textit{VividDreamer} achieves superior generation quality with the same optimization time, presenting a noteworthy training efficiency.
It also demonstrates that \textit{VividDreamer} significantly addresses the bottleneck in existing text-to-3d generation methods, proving the correctness and effectiveness of our motivation and model design.

\subsubsection{User Study}
We conduct a user study to provide a comprehensive evaluation of text-to-3D generation quality.
Specifically, we randomly select 50 prompts from a total of 414 ones for evaluation, and generate 3D objects using 5 different methods (including our 10- and 30-minute versions) with each prompt.
For fair comparisons, we anonymize all the methods, and exhibit each 3D asset by its 360$^\circ$ rendered video.
Users are invited to fill out the questionnaire online, and asked to rank the 3D objects based on the fidelity and the degree of alignment with the given text prompt.
\Cref{tab:us} reports the results summarized from 98 questionnaires on the internet, where we evaluate users' preferences using the average ranking.
As shown, our \textit{VividDreamer} achieves the highest average ranking, demonstrating that users consistently favored the 3D objects generated by our framework. 
This also demonstrates that \textit{VividDreamer} achieves a high training efficiency, only 10 minutes of training on a single A100 GPU can outperform the competitors.

\subsection{Ablation Study}
In this subsection, we carry out ablation studies to investigate the effectiveness of the proposed modules.

\subsubsection{Different Objectives}
We evaluate the visual quality of our framework based on different objectives: SDS, ISM, and our \textit{PCDS}.
Specifically, for each objective, we incorporate it into our framework and conduct training of 3000 iterations.
All the hyperparameters and model settings are consistent.
During training, we gradually increase the sampling steps of our \textit{PCDS} from 1 to 3.
As shown in \cref{fig:loss_comp}, \textit{PCDS} achieves better generation quality than SDS and ISM, producing high-fidelity and semantically-inconsistent 3D assets.
This also proves that our \textit{PCDS} can alleviate the color deviation and Janus problem caused by the inaccurate gradient estimation.

\subsubsection{Different Optimization Strategies}
\begin{figure}[!t]
  \centering
  \includegraphics[width=0.48\textwidth]{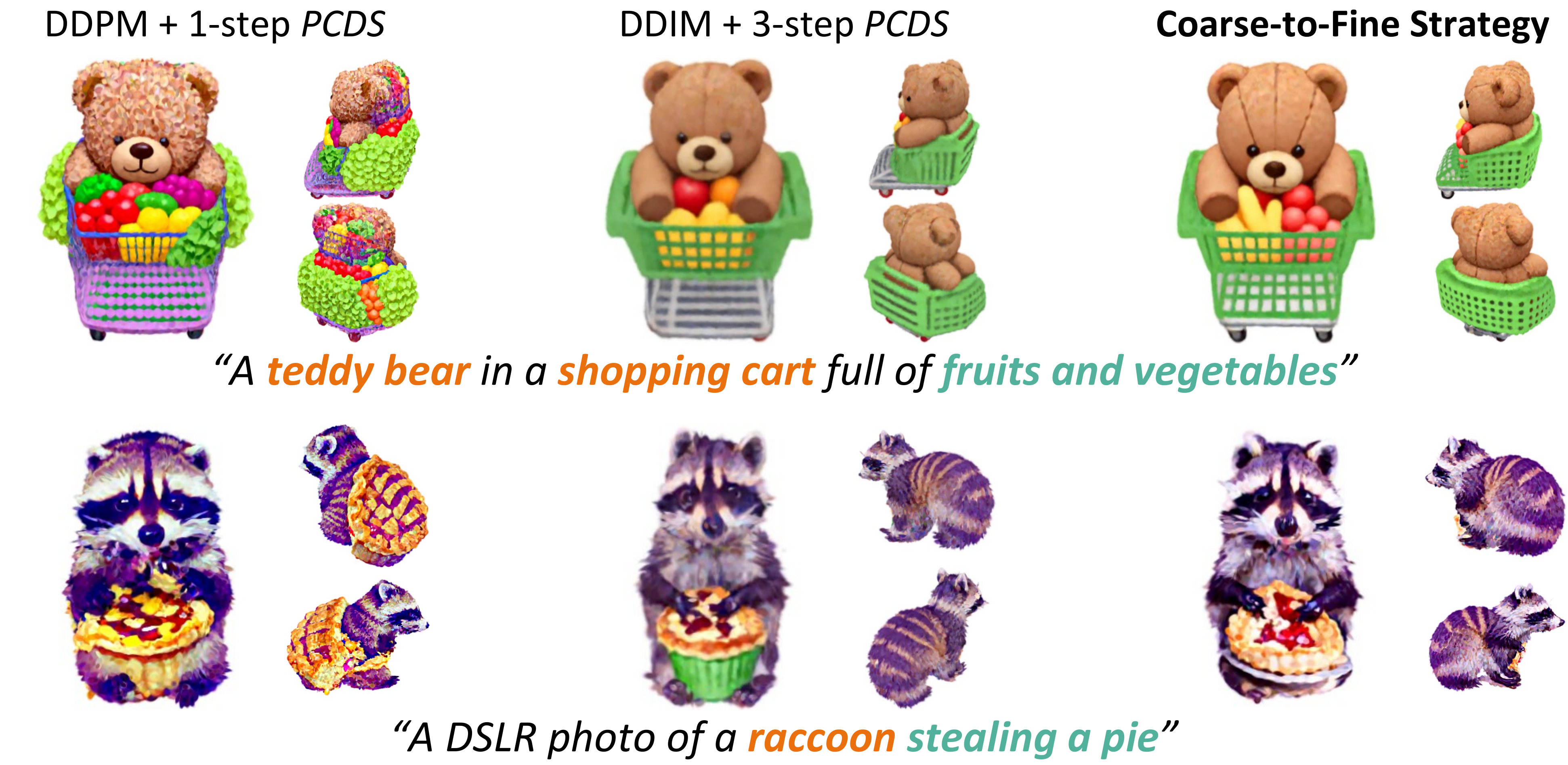}
  \caption{
    Visual comparisons between different ablations of our framework. The results are generated under 10 minutes of training on a single A100 GPU.
    Visually, the proposed coarse-to-fine optimization strategy acquires superior generation quality to the competitors in the same training time, achieving better training efficiency.}
  \label{fig:abctf}
\end{figure}

In \cref{fig:abctf}, we evaluate the training efficiency of the following optimization strategies: \textit{only} 1-step \textit{PCDS}, \textit{only} 3-step \textit{PCDS}, and our coarse-to-fine strategy, \textit{i.e.,} gradually increasing the sampling steps of \textit{PCDS} from 1 to 3.
With the same training time (10 minutes), our coarse-to-fine results achieve higher generation quality, producing 3D-consistent structures and finer details, which proves the effectiveness of our designs.

\begin{table}[!t]
  \caption{We survey the average ranking of users' preference on 50 sets of text-to-3D generation results produced by the state-of-the-art and our 10- and 30-minute versions, respectively. Our results are preferred by most users. The training time is evaluated on a single A100 GPU.}
  \label{tab:us}
  \centering
  \begin{tabular}{lcc}
  \toprule
  Methods                                      & \textit{Avg.} Rank $\downarrow$ & Training Time $\downarrow$\\\midrule
  DreamGaussian \cite{tang2023dreamgaussian}   & 4.74                   & 3 \textit{mins} \\
  GaussianDreamer \cite{yi2024gaussiandreamer} & 3.41                   & 9 \textit{mins} \\
  LucidDreamer \cite{liang2024luciddreamer}    & 2.65                   & 45 \textit{mins} \\
  \rowcolor{MyGray}
  \textbf{VividDreamer}$_{10}$ (Ours)      & \textbf{2.26}          & 10 \textit{mins} \\
  \rowcolor{MyGray}
  \textbf{VividDreamer}$_{30}$ (Ours)      & \textbf{1.43}          & 30 \textit{mins} \\
  \bottomrule
  \end{tabular}
\end{table}

\begin{figure*}[!t]
  \centering
  \includegraphics[width=\textwidth]{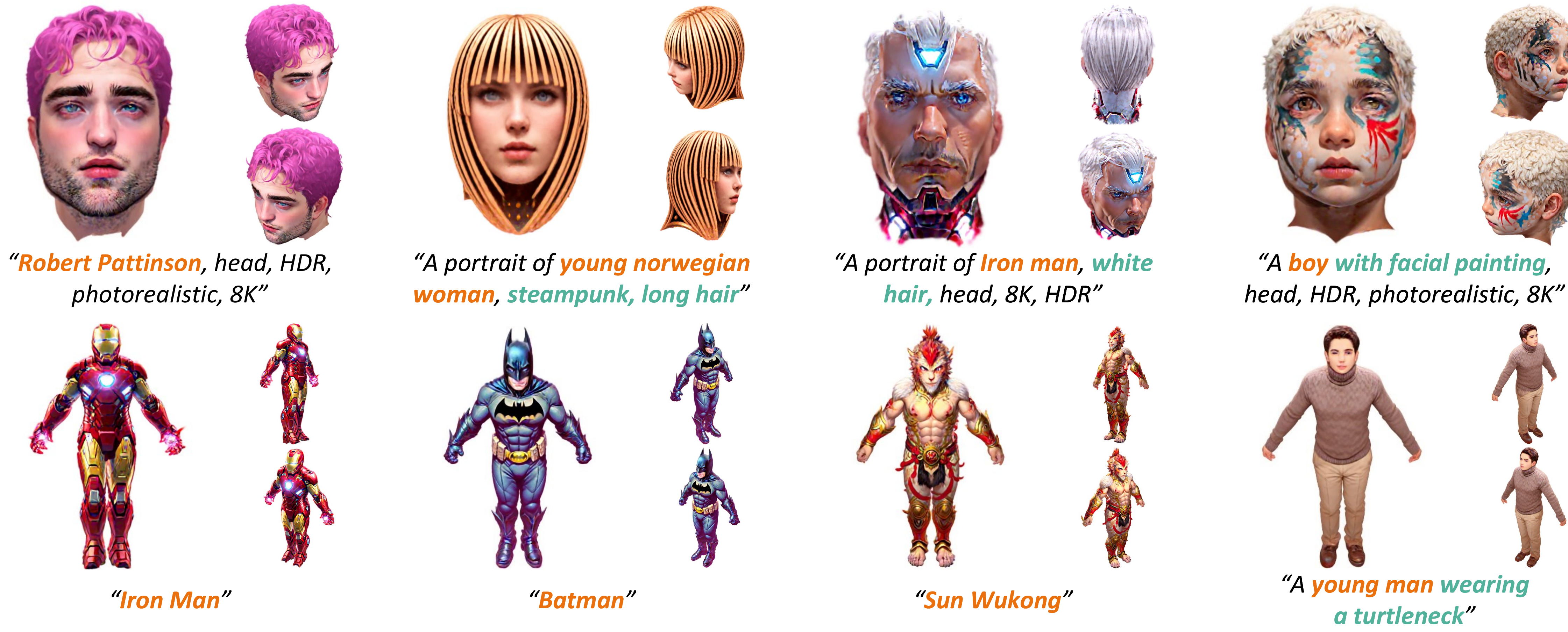}
  \caption{
    Visual examples of 3D portraits and avatars.
    As shown, our approach can create high-quality 3D avatars and portraits within 30 minutes of training on a single A100 GPU.
  }
  \label{fig:avatar}
\end{figure*}

\begin{figure*}[!t]
  \centering
  \includegraphics[width=\textwidth]{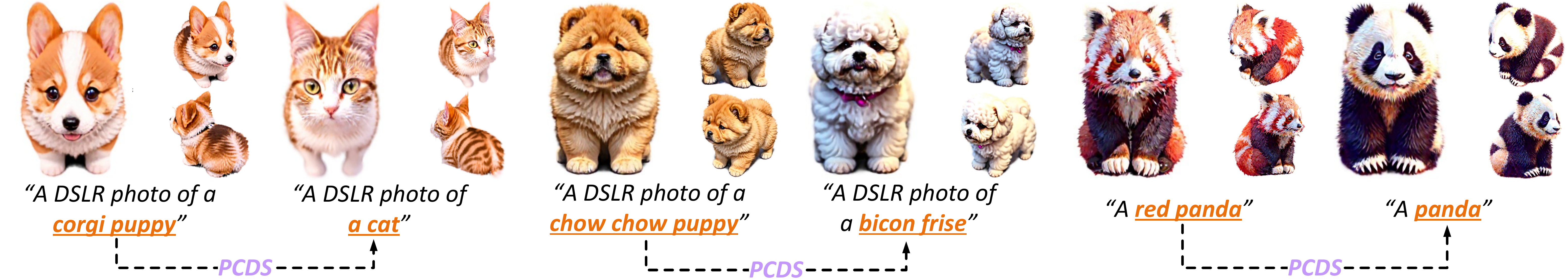}
  \caption{
    Visual examples of editing the given 3D assets with user-guided prompts using the proposed \textit{PCDS}.
    As shown, our approach can create user-desired 3D objects while maintaining the basic structure of the prototypes.}
  \label{fig:editing}
\end{figure*}

\section{Applications}\label{sec:app}
In this section, we further explore the effectiveness of our work on other 3D generative applications, such as zero-shot 3D avater and portrait generation, and zero-shot 3D editing.

\subsection{Zero-shot 3D Avatar \& Portrait Generation}
We expand our \textit{VividDreamer} to generate 3D avatars and portraits from text descriptions.
To generate 3D avatars, we utilize the Skinned Multi-Person Linear Model (SMPL) \cite{SMPL} to generate the point cloud of the human body as a geometry prior to 3DGS initialization.
We also employ ControlNet \cite{zhang2023adding} conditioned on depth maps to offer more robust supervision during training.
For 3D portrait creation, 3DGS is initialized by the point cloud of ``head'' provided from Point-E \cite{point-e}.
As shown in \cref{fig:avatar}, only following such a simple process, \textit{VividDreamer} can generate high-fidelity 3D avatars (row 1) and portraits (rows 2 and 3) that consistent with text prompts.

\subsection{Zero-shot 3D Editing}
In addition to text-to-3D generation tasks, our \textit{PCDS} can also be simply extended to 3D editing tasks.
Because \textit{PCDS} can acquire updating direction that consistent with text prompts, it enables editing the textures of an arbitrary 3D asset based on a user-guided text prompt in a zero-shot manner.
Specifically, given a 3D asset and a text prompt, we first transform the 3D asset into a point cloud to initialize 3DGS.
Then, we employ the depth map-prior ControlNet \cite{zhang2023adding} to guide the optimization.
As shown in \cref{fig:editing}, the edited results show high-quality details consistent with the given text prompts, while maintaining the basic structure of original 3D objects.

\section{Conclusions}\label{sec:con}
In this work, we present VividDreamer, a text-to-3D generation framework that significantly improves the efficiency of 3D content creation. 
We design an efficient Pose-dependent Consistency Distillation Sampling (PCDS) objective, and propose a coarse-to-fine optimization framework, enabling high-fidelity 3D object creation in a short time. 
Thanks to the proposed modules, our approach can first create ready-to-use 3D assets within 10 minutes, and then produce high-fidelity 3D structures and fine-grained details within 30 minutes. 
Extensive experiments demonstrate that our \textit{VividDreamer} favorably outperforms the state-of-the-art in generation quality and training efficiency. 
Its outstanding performance paves the way for a wide range of 3D generative applications, such as text-to-3D editing and zero-shot avatar creation, making it possible to facilitate numerous applications in the real world.


%
%
%
%

\bibliographystyle{IEEEtran}
\bibliography{egbib}

\vfill

\end{document}